\documentclass[conference]{IEEEtran}
\IEEEoverridecommandlockouts
\usepackage{cite}
\usepackage[normalem]{ulem}
\usepackage{graphicx}
\usepackage{subcaption}
\usepackage{float}
\usepackage{multirow}
\usepackage{amsmath,amssymb,amsfonts}
\usepackage[ruled,linesnumbered]{algorithm2e}
\usepackage{algorithmic}
\usepackage{textcomp}
\usepackage{xcolor}
\usepackage{marvosym}

\usepackage{hyperref}
\hypersetup{
    colorlinks=true, 
    linkcolor=blue, 
    filecolor=magenta, 
    urlcolor=cyan, 
    citecolor=green 
}
\newcommand\blfootnote[1]{%
  \begingroup
  \renewcommand\thefootnote{}\footnote{#1}%
  \addtocounter{footnote}{-1}%
  \endgroup
}

\usepackage{tcolorbox}

\def\BibTeX{{\rm B\kern-.05em{\sc i\kern-.025em b}\kern-.08em
    T\kern-.1667em\lower.7ex\hbox{E}\kern-.125emX}}

\title{Goldfish: An Efficient Federated Unlearning Framework}

\author{
\IEEEauthorblockN{Houzhe Wang\textsuperscript{1,2}, Xiaojie Zhu\textsuperscript{3(\Letter)},  Chi Chen\textsuperscript{1,2}, Paulo Esteves-Veríssimo\textsuperscript{3}}
\IEEEauthorblockA{
\textit{School of Cyber Security, University of Chinese Academy of Sciences, Beijing, China} \\
\textit{Institute of Information Engineering, Chinese Academy of Sciences, Beijing, China} \\
\textit{King Abdullah University of Science and Technology, Thuwal, Kingdom of Saudi Arabia  } \\
\{wanghouzhe, chenchi\}@iie.ac.cn \\
\{xiaojie.zhu, paulo.verissimo\}@kaust.edu.sa
}
\thanks{(\Letter): Corresponding author}
}
\begin{document}
\maketitle
\thispagestyle{plain}

\begin{abstract}
With recent legislation on the right to be forgotten,  machine unlearning has emerged as a crucial research area.
It facilitates the removal of a user's data from federated trained machine learning models without the necessity for retraining from scratch.
However, current machine unlearning algorithms are confronted with challenges of efficiency and validity.

To address the above issues, we propose a new framework, named Goldfish. 
It comprises four modules: basic model, loss function, optimization, and extension.
To address the challenge of low validity in existing machine unlearning algorithms, we propose a novel loss function.
It takes into account the loss arising from the discrepancy between predictions and actual labels in the remaining dataset.
Simultaneously, it takes into consideration the bias of predicted results on the removed dataset.
Moreover, it accounts for the confidence level of predicted results.
Additionally, to enhance efficiency, we adopt knowledge a distillation technique in the basic model and introduce an optimization module that encompasses the early termination mechanism guided by empirical risk and the data partition mechanism.
Furthermore, to bolster the robustness of the aggregated model, we propose an extension module that incorporates a mechanism using adaptive distillation temperature to address the heterogeneity of user local data and a mechanism using adaptive weight to handle the variety in the quality of uploaded models. Finally, we conduct comprehensive experiments to illustrate the effectiveness of proposed approach.

\blfootnote{Our code is available at \href{https://github.com/xiao-jian-zi/MU-Goldfish-An-Efficient-Federated-Unlearning-Framework}{https://github.com/xiao-jian-zi/MU-Goldfish-An-Efficient-Federated-Unlearning-Framework.}}

\end{abstract}


\section{Introduction}

Federated learning is an advanced distributed machine learning paradigm 
that enables multiple clients to collaboratively train a shared global model 
without the need to share their local data \cite{konevcny2016federated} \cite{mcmahan2017communication}. 
This approach effectively addresses a key challenge in traditional machine learning: enabling model training in the absence of centralized storage and processing for datasets.
Federated learning effectively addresses this limitation through distributed training, empowering data holders to retain ownership and control of their data while actively contributing to the training of a global model. In this framework, participants are only required to upload the updated parameters of their local models to the central server. The server then integrates these updates to construct an improved global model. This process serves to safeguard client data privacy.

However, within the federated learning framework, users may request the removal of their contribution from the trained global model. Moreover, recent regulations such as the European Union's General Data Protection Regulation (GDPR) \cite{regulation2018general} and the California Consumer Privacy Act (CCPA) \cite{pardau2018california} empower individuals with the right to demand the deletion of their private data from any part of the system within a reasonable time frame. Furthermore, even if the original data was never shared, the global machine learning model could still glean information about the clients \cite{nasr2018comprehensive} \cite{song2020analyzing}. Predictions made by the global model might potentially leak client information \cite{salem2018ml}\cite{bagdasaryan2020backdoor}. Therefore, there is a compelling need for a method to eliminate a client's contribution from the trained global model.


To ensure that a global model forgets the contributions of a specific client, one straightforward approach is to retrain the model from scratch after removing the target user's data.
 Alternatively, another method is to remove user information from the parameters of the trained model, while preserving the utility of the model \cite{bourtoule2021machine}.
 
In the context of managing large datasets and complex models, retraining approaches can be impractical due to high time and energy costs, computational expenses, and scalability challenges.
There is a growing interest in developing cost-effective machine unlearning algorithms to mitigate the impact of deleted data from trained models.
To measure the effectiveness of an unlearning algorithm in eradicating specified information, 
Ginart \textit{et al.} \cite{ginart2019making} introduced a metric similar to ($\varepsilon$, $\delta$)-differential privacy (DP). This metric serves as proof of indistinguishability between the output of the unlearning algorithm and the newly retrained output without the deleted records.
Building on this concept, several certifications of forgetting have been introduced, aiming to validate the efficacy of various forgetting mechanisms \cite{izzo2021approximate,guo2019certified,sekhari2021remember,neel2021descent,ullah2021machine}.

However,  researchers have demonstrated the weaknesses in current machine unlearning algorithms.
Chourasia \textit{et al.} \cite{truedata} recognizes the interdependence between training data and the model, emphasizing that the data used in the training process imbues specific patterns within the model. In alignment with  Chourasia \textit{et al.}, Gupta \textit{et al.} \cite{gupta2021adaptive}  contend that there are inherent flaws in the deletion certification approach of previous unlearning works, which rely on the indistinguishability between unlearned and retrained models.


Moreover, the existing unlearning algorithms  \cite{ginart2019making,izzo2021approximate,guo2019certified}  
enhance the efficiency of machine unlearning algorithms by relaxing the requirements for validity. {Here, 
`validity' refers to the effectiveness of the unlearning process in ensuring that the model no longer retains information about the deleted data.}
Furthermore, most of these algorithms necessitate access to users' historical model updates or gradient information to enhance efficiency, which is impermissible within the framework of federated learning. 
Recent studies \cite{chen2021machine,zhu2019deep,huang2021evaluating} have demonstrated that attackers, such as a malicious central server, can exploit clients' local gradients to mount attacks that reconstruct private training samples of the clients.


In this paper, we aim to overcome the aforementioned shortcomings.
First, to address the challenge of low validity in existing machine unlearning algorithms, we propose a novel loss function.
It takes into account the loss arising from the discrepancy between predictions and actual labels in the remaining dataset.
Simultaneously, it takes into consideration the bias of predicted results on the removed dataset.
Moreover, it accounts for the confidence level of predicted results.
Additionally, to enhance efficiency, we adopt knowledge distillation technique in basic model and introduce an optimization module that encompasses the early termination mechanism guided by empirical risk and the data partition mechanism.
Furthermore, to bolster the robustness of the aggregated model, we propose an extension module that incorporates a mechanism using adaptive distillation temperature to address the heterogeneity of user local data and a mechanism using adaptive weight to handle the variety in the quality of uploaded models.

Overall, the main contributions of this paper are summarized as follows. 

\begin{itemize}
    \item We introduce a novel paradigm for designing a machine unlearning algorithm, named Goldfish. It comprises four modules: basic model, loss function, optimization, and extension.
    
    \item We propose a novel design for the loss function that incorporates the discrepancy between predictions and actual labels in the remaining dataset, the bias of predicted results on the removed dataset, and the confidence level of predicted results.

    \item We propose an optimization module that incorporates the early termination mechanism guided by empirical risk and the data partition mechanism. Additionally, we introduce an extension module that includes a mechanism using adaptive distillation temperature to address the heterogeneity of user local data and a mechanism using adaptive weight to handle the variety in the quality of uploaded models.

    \item We have established the foundational framework of Goldfish and conducted comprehensive experiments on publicly known datasets to assess its effectiveness.
    The experimental results demonstrate that the proposed approach can effectively resist backdoor attacks while exhibiting better efficiency and accuracy compared to the state-of-the-art methods.
\end{itemize}

\section{Related Work}

Federated unlearning encompasses two primary approaches: retraining-based and model update adjustment-based methods \cite{liu2022right}. 
Retraining-based methods require substantial retraining of the global model using client data, leading to significant time and resource consumption.
In contrast, model update adjustment methods utilize parameter updates from the client to revise the model, eliminating the need for complete retraining.
While update adjustment techniques are generally faster and more resource-efficient, they demand the retention of additional information and meticulous calibration and aggregation to ensure efficacy.


\subsection{Retraining-based approach}
Numerous research efforts have been undertaken to expedite the retraining process and minimize time overhead.
 Liu \textit{et al.} \cite{liu2022right} leverages the first-order Taylor expansion approximation technique to customize a rapid retraining algorithm based on diagonal experience FIM. 
On the contrary, Yuan \textit{et al.} \cite{yuan2023federated} presents a federated forgetting framework. This framework empowers clients to request data deletion, prompting the server to retrain the global model based on these removal requests.

Bourtoule \textit{et al.} \cite{bourtoule2021machine} introduced SISA training as an approach to alleviate computational costs associated with forgetting. 
This method strategically limits the influence scope of data points during training by employing techniques such as data sharding and slicing.

\subsection{Model update adjustment-based approach}

Much research has been conducted with the aim of enhancing efficiency and effectiveness in the model update adjustment-based approach.
Zhang \textit{et al.} \cite{zhang2023fedrecovery}  eliminate client influence by extracting the weighted sum of gradient residuals from the global model and customizing Gaussian noise. 
This process is designed to achieve statistical indistinguishability between unlearned and retrained models.
Liu \textit{et al.} \cite{liu2021federaser} reconstruct the forgotten model using parameter updates stored on the server, introducing a novel calibration method to adjust client updates. 
This innovative approach aims to enhance forgetting speed while preserving model performance.
In addition, Baumhauer \textit{et al.} \cite{baumhauer2022machine} and Thudi \textit{et al.} \cite{thudi2022necessity} emphasize the pursuit of higher efficiency in machine unlearning by relaxing requirements for both effectiveness and provability.
Izzo \textit{et al.} \cite{izzo2021approximate}, Neel \textit{et al.} \cite{neel2021descent}, and Wu \textit{et al.} \cite{wu2020deltagrad} explore techniques for the server to effectively approximate gradients during the unlearning process by leveraging historical gradients and model weights.
Chourasia \textit{et al.} \cite{chourasia2023forget} emphasize the significance of robustness when addressing data deletion scenarios.

{
The proposed Goldfish framework presents a general approach for realizing federated unlearning, comprising four key modules: basic model, loss function, optimization, and extension. The Goldfish framework accommodates the implementation of both retraining-based methods and model update adjustment-based methods.
In addition, we introduce a more effective loss function designed to achieve a balance between accuracy and validity in data deletion scenarios. 
Furthermore, the framework incorporates an adaptive distillation temperature mechanism to address client-side data heterogeneity. Additionally, we propose an adaptive weight mechanism to effectively manage variations in model quality.
Finally, to boost the efficiency of the retraining process, we introduce a data sharding mechanism, complemented by an early termination mechanism guided by empirical risk. 
}

\section{Proposed Framework: Goldfish}
In this section, we introduce the proposed framework: Goldfish. 
We first demonstrate the overview of Goldfish and then illustrate its details. 
Prior to introducing the scheme, we illustrate the commonly used notation in Table \ref{tab:notation}. 
\begin{table}[]
    \centering
     \caption{Frequently used notations.}
    \label{tab:notation}
    \begin{tabular}{|c|l|}
    \hline
     $D_c$ & the dataset belonging to client $c$  \\
    \hline  
     $D_f^c$  &  the   removed dataset of client $c$  \\ 
    \hline 
     $D_r^c$   & the remaining dataset of client $c$ \\ 
    \hline 
    $\alpha$ &  the total number of classes of labels  \\ 
    \hline 
    $\tau$ & the total number of shards of user local data \\ 
    \hline 
    $t$ & the number of epochs \\
    \hline 
    $C$ & the total number of clients \\ 
    \hline 
    $\omega$ & global model parameter \\ 
    \hline 
    $\omega_c$ & local model parameter \\ 
    \hline 
    \end{tabular}
\end{table}

\subsection{Overview of Goldfish}
\begin{figure}
    \centering
    \includegraphics[width=0.49\textwidth]{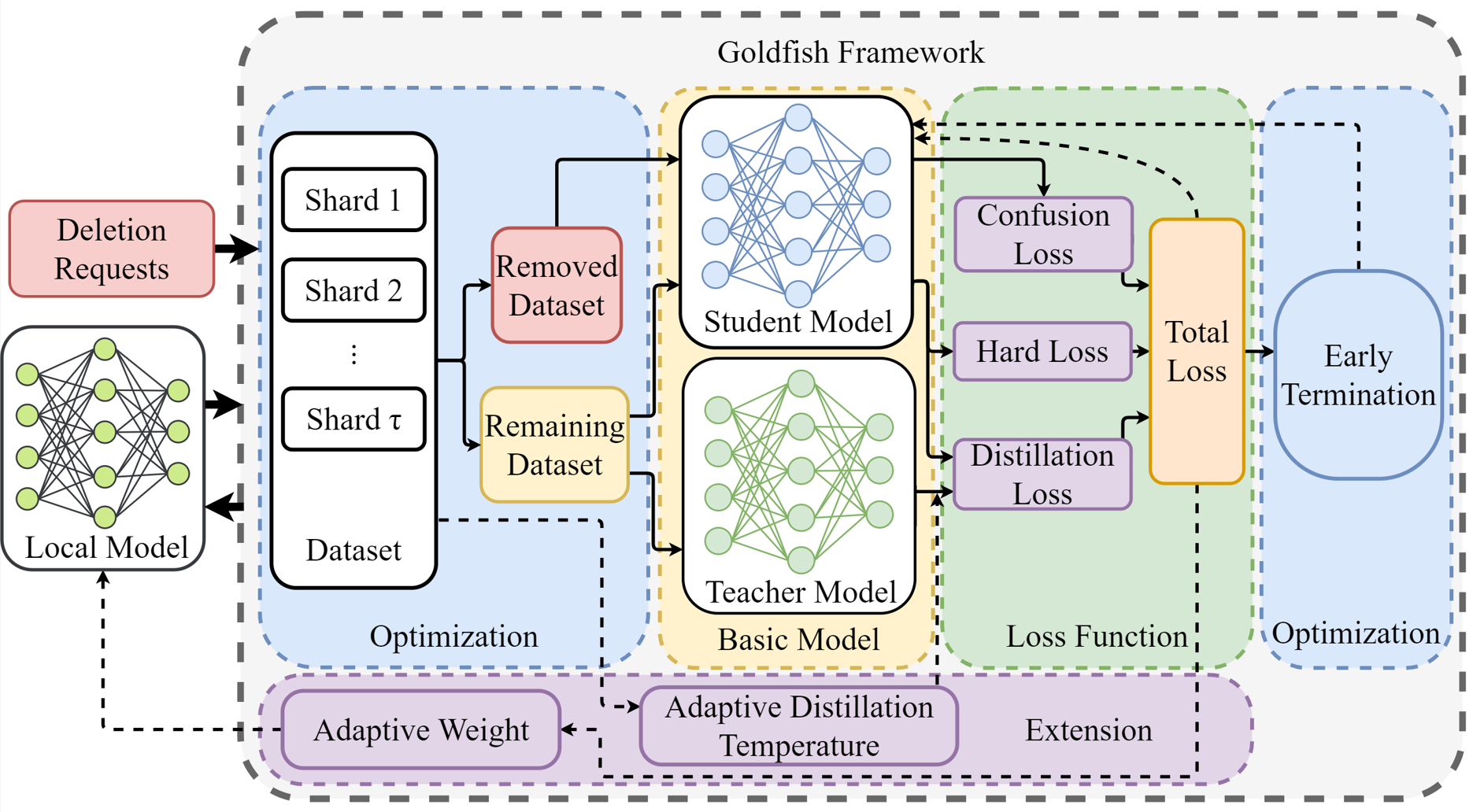}
    \caption{Goldfish Framework. It consists of four modules: basic model, loss function, optimization, and extension.
    }
    \label{fig:famework}
\end{figure}

As shown in Fig.\ref{fig:famework}, Goldfish specializes in four modules: basic model, loss function, optimization, and extension.

The initial module of the Goldfish,  basic model,  is crafted for model selection in the context of federated unlearning. The second module is dedicated to the implementation of a personalized loss function. The third module is engineered to execute an optimization algorithm aimed at enhancing efficiency. The final module is tailored to address any additional requirements.

The underlying principle guiding this design is rooted in the characteristics of federated unlearning.
In federated learning, as the first step,  we need to select the initial model to 
implement the unlearning process.  In particular, the trace generated from the data that is requested to be removed should be eliminated from the global model. 
To enhance the effectiveness of the selected model, the design of the loss function is pivotal. Loss functions provide a precise metric for assessing a model's performance by quantifying the disparity between predictions and actual results.

In addition, given the sporadic nature of data removal requests, ensuring the efficient execution of the unlearning process is imperative. To adapt to the evolution of efficient approaches, an optimization module is provided.
Lastly, we offer an extension module to achieve compatibility.
Additional requirements can be incorporated into this module. 
For instance, the approach to address client heterogeneity can be implemented within this module.

\begin{algorithm}
    \caption{Goldfish Algorithm}
    \label{alg:goldfish}
    \SetAlgoLined
    \SetKwInOut{Input}{Input}
    \SetKwInOut{Output}{Output}

    \Input{local training dataset $D_c$, initialized model parameter $\omega_0$, learning rate $\mu$, deletion request}
    \Output{unlearned global model}
    \SetKwFunction{Goldfish}{Goldfish}
    \SetKwProg{Server}{Server}{:}{}
    \SetKwProg{Client}{Client}{:}{}
    \SetKwProg{Procedure}{Procedure}{:}{}
    \Procedure{Efficient Federated Unlearning Framework}{}{
        Reinitialize global model and distribute to all clients $c$: $\omega_c^0 \leftarrow \omega_0$\;
        \For{$t = 0, 1, \ldots, N$}{
            \ForEach{client $c$ in parallel}{
                \If{no deletion request}{
                    LocalTraining($\omega_c^t$, $D_c$)\;
                }
                \Else{
                    \ForEach{unlearned client $c$ in parallel}{
                        $D_f^c\leftarrow$ deleted data\;
                        $D_r^c \leftarrow D_c / D_f^c$\;
                        Reinitialize global model and distribute to all clients $c$: $\omega_c^t \leftarrow \omega_0$\;
                        $\omega_c^{t+1} \leftarrow $\Goldfish{$\omega_c^t$, $\omega_{t-1}$, $D_f^c$, $D_r^c$}\;
                    }
                    \ForEach{normal client $c$ in parallel}{
                        $\omega_c^{t+1} \leftarrow $\Goldfish{$\omega_c^t$, $\omega_{t-1}$, $D_f^c$, $D_r^c$}\;
                    }
                }
            }
            \Server{}{
                Update global model parameters $\omega_{t+1}$\;
            }
        }
        \Return $\omega_{t+1}$\;
    }

    \vspace{5mm}

    \Procedure{\Goldfish{$\omega_c^t$, $\omega_t$, $D_f^c$, $D_r^c$}}{
        Initialize teacher model $M_T$ parameters: $\omega_T \leftarrow \omega_t$\;
        Initialize student model $M_S$ parameters: $\omega_S \leftarrow \omega_c^t$\;
        \ForEach{local epoch $i=0,1,2,\ldots,n$}{
                \If{$D_f^c$ is not empty}{
                    $\omega_c^t(i+1) \leftarrow \omega_c^t(i) - \mu\nabla l(M_T,M_S,D_f^c,D_r^c)$\;
                }
                \Else{
                    $\omega_c^t(i+1) \leftarrow \omega_c^t(i) - \mu\nabla l(M_T,M_S,D_r^c)$\;
                }
        }
        \Return $\omega_c^t(i+1)$\;
    }

    \vspace{5mm}

    \Procedure{LocalTraining($\omega_c^t$, $D_c$)}{
        \ForEach{local epoch $i=0,1,2,\ldots,n$}{
            $\omega_c^t(i+1) \leftarrow \omega_c^t(i) - \mu\nabla l(\omega_c^t,D_c)$\;
        }
        
        \Return $\omega_c^t(i+1)$\;
    }

\end{algorithm}

\subsection{Details of Modules}
In this section, we delve into the specifics of each module within Goldfish, and its detailed description of the algorithm is presented in Algorithm \ref{alg:goldfish}. 

\noindent\textbf{Basic model}. 
In Goldfish, the basic model achieves fast retraining through knowledge distillation. 
In particular, we utilize a teacher model $M_T$ and a student model $M_S$ to implement the retraining process on a user's local dataset $D_c$. 
The local dataset $D_c$ consists of two parts: $D_f^c$ and $D_r^c$. 
$D_f^c$ represents the data that needs to be removed while $D_r^c$ denotes the remaining data. 
The student model $M_S$ can be initialized without any knowledge of $D_c$, while the global model serves as the teacher model $M_T$. 
The teacher model $M_T$ encompasses the knowledge acquired from both datasets $D_r^c$ and $D_f^c$. 
Our approach is designed to enable the student model $M_S$ to selectively learn from the teacher model, preserving exclusively the knowledge associated with the remaining dataset $D_r^c$, thereby accomplishing the forgetting of $D_f^c$.
During the training process, we ensure that the knowledge transfer between the teacher model $M_T$ and the student model $M_S$ occurs exclusively on dataset $D_r^c$, fundamentally preventing the student model from learning the removed data $D_f^c$.
Moreover, we can employ the removed dataset $D_f^c$ to evaluate the output of the student model, ensuring that predictions on the removed data are not biased. 


\noindent\textbf{Loss function}. The proposed loss function is an amalgamation of hard loss, confusion loss, and distillation loss.

\textit{Hard loss}. 
The discrepancy between the model's prediction and the actual label is named as ``hard loss". 
The hard loss of the unlearning process on $D_c$ can be split into two aspects. 
The first aspect is associated with the remaining dataset $D_r^c$.  
During the training, the label prediction of $D_r^c$ may deviate from the actual label. 
The total loss caused by the $D_r^c$ is denoted as $L_{r}$.
Another aspect is related to the removed data $D_f^c$.  
The total loss caused by $D_f^c$ is 
denoted as $L_{f}$.
Finally, the hard loss is represented by Equation \ref{eq:hard_loss}.
In particular, in  multi-class classification, $L_r$ is computed by 
$-\sum_{x_i \in D_r^c} y_i \log (M_s(x_i))$, where 
$M_s(x_i)$ represents the prediction made by $M_s$ on the data {$x_i$} from $D_r^c$, {$y_i$ represents the true label corresponding to $x_i$.}
 Similarly, $L_f$ is computed by 
$-\sum_{x_i \in D_f^c} y_i \log (M_s(x_i))$, where 
$M_s(x_i)$ represents the prediction made by $M_f$ on the data $x_i$ from $D_f^c$. 
In the paper, we consider the size of $D_r^c$ is much larger than $D_f^c$ ($|D_r^c| >> |D_f^c|$). 
Without losing generality,  the minimization of $L_h$ promotes  $L_f$ while suppressing $L_r$.

\begin{equation}
\label{eq:hard_loss}
    L_{h}=L_{r} - L_f
\end{equation}



\textit{Confusion loss}. 
The confusion loss is designed to reduce the bias of the predicted results of $M_S$ on $D_f^c$. 
We set the Equation \ref{eq:confusion_loss} to evaluate the bias level of the predicted results, 
\begin{equation}
\label{eq:confusion_loss}
    L_{c}=\frac{1}{\left|D_f^c\right|}\sum_{x_j\in D_f^c}\sqrt{D\left(M_S\left(x_j\right)\right)}
\end{equation}
 where $M_S(x_j)$ represents the predicted vector of the student model for the samples in the removed dataset $D_f^c$ and $D(M_S(x_j))$ denotes the variance of the predicted vector.
 This confidence vector contains the predicted probabilities of a sample belonging to various classes.
Variance measures the dispersion between predicted results, and we aim at minimizing it, ensuring that the confidence in predictions made by $M_S$ for each sample in $D_f^c$ is close. 

\textit{Distillation loss}. 
Knowledge distillation model utilizes the predicted results generated by the teacher model as labels for the student network. 
The predicted results of the teacher model are transformed into prediction confidences through the softmax function. 
The confidence level $P_{x_i}^T$ of a sample $x_i$ ($x_i \in D_r^c$) for teacher model $M_T$ is computed as Equation \ref{eq:dprobT}, 
\begin{equation}
\label{eq:dprobT}
    P_{x_i}^T=\frac{exp{\left(v_i/T\right)}}{\mathrm{\Sigma_{j=1}^C}exp{\left(v_j/T\right)}}
\end{equation}
where $T$ represents the distillation temperature, 
$v_j$ represents the confidence score of the teacher model $M_T$ predicting $x_i$ belonging to class $j$, 
and $v_i$ represents the confidence score of the teacher model $M_T$ predicting $x_i$ belonging to the class $i$ ($i \in [1, \alpha]$) correctly. 
From  Equation \ref{eq:dprobT}, we can learn that higher distillation temperatures can make the predicted probabilities generated by the teacher model smoother.
Using the same methodology, we define the confidence level $P_{x_i}^S$ of a sample $x_i$ for student model $M_S$ as Equation \ref{eq:dprobS}, 
\begin{equation}
\label{eq:dprobS}
    P_{x_i}^S=\frac{exp{\left(z_i/T\right)}}{\mathrm{\Sigma_{j=1}^C}exp{\left(z_j/T\right)}}
\end{equation}
where $z_j$ represents the confidence score of the student model $M_S$ predicting $x_i$ belonging to class $j$, 
and $v_i$ represents the confidence score of the student model $M_S$ predicting $x_i$ belonging to the class $i$ ($i \in [1, \alpha]$) correctly.

Finally, the distillation loss is defined based on the confidence level, as formalized by Equation \ref{eq:dsl}.
The equation shows that 
the greater the disparity between the predicted distributions of the teacher model and the student model, the larger the loss.

\begin{equation}
    \label{eq:dsl}
    L_{d}=-\sum_{x_i\in D_r^c} P_{x_i}^T \log P_{x_i}^S  
\end{equation}

\textit{Final expression of loss function}. The final loss function is defined as follows:
\begin{equation}
    L=L_{h}+{\mu_{c}L}_{c}+{\mu_{d }L}_{d}
\end{equation}
where the weight factors $\mu_{c}$ and $\mu_{d}$ can be adjusted to balance the importance of different objectives. Assigning a larger weight to the confusion loss can reduce the bias of model $M_S$ on $D_f^c$. 
The hard loss ensures that the model learns from the remaining dataset  $D_r^c$, 
while the distillation loss improves the generalization of the model. 


\noindent\textbf{Optimization}. We introduce two approaches to improve the efficiency of retraining.  
The first method is implemented through terminating the training process earlier based on convergence speed.
The second is to divide the data into many small portions. The retraining process only needs to be conducted on those portions that include deleted data.

\textit{Early termination guided by empirical risk}. 
During the training process, the loss value output by the loss function is used to measure the convergence of the model. 
To decide the condition of stopping training, we introduce excess empirical risk to determine whether the loss value of the student model during the training process is close to that of the global model from the previous epoch.
The excess empirical risk is defined in Equation \ref{eq:eem}, where $\omega^{t-1}$ represents the global model  parameters of $t-1$ epoch of training, and
 $w_c^t(i)$ represents the $i$th ($i \in [0,n]$) epoch of local training corresponding to the $t$ epoch of global training, and $n$ stands for the number of epochs of local training. 

\begin{equation}
\label{eq:eem}
    err\left(\omega_c^t,\omega^{t-1}\right)=\left|\frac{1}{n}\sum_{i=0}^{n}L\left(\omega_c^t\left(i\right)\right)-L\left(\omega^{t-1}\right)\right|
\end{equation}


During the training phase, when the loss value of the student model decreases to within a certain range of the lowest loss of the pre-trained model, i.e., $err\left(w_c^t,w^{t-1}\right)\le\delta$, where $\delta$ is a specified threshold value,
 the training process terminated. 

\begin{figure}
    \centering
    \includegraphics[width=0.4\textwidth]{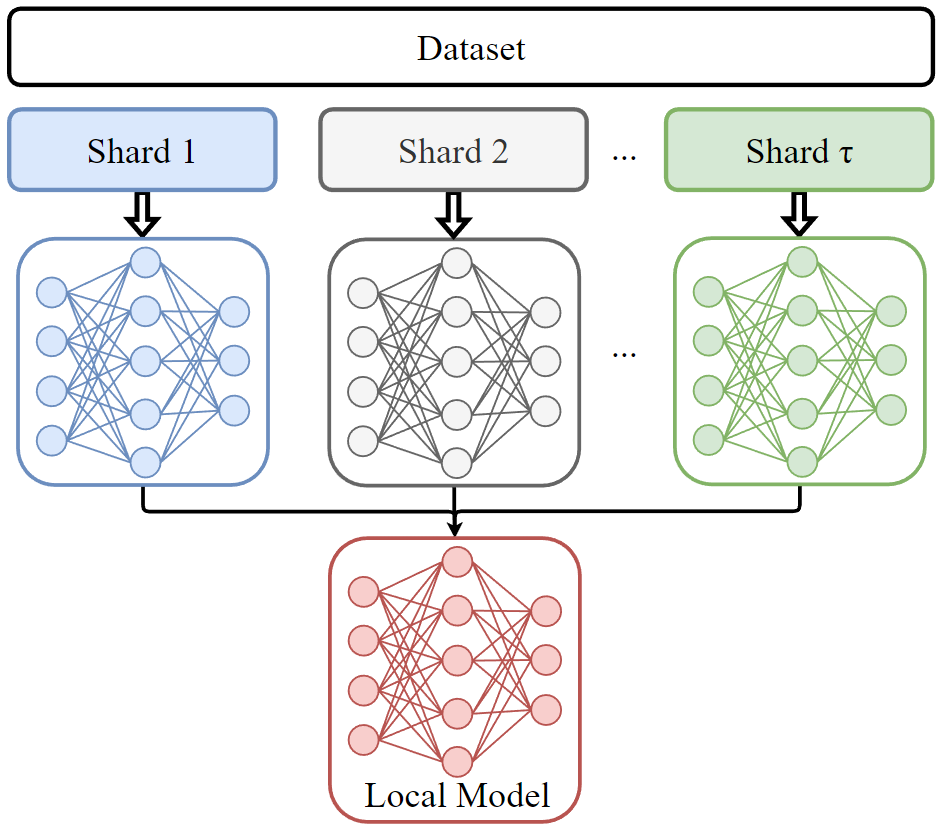}
    \caption{Data Sharding Diagram. Each dataset is partitioned into data shards. 
Each shard has a model and the final output is the aggregation of models from these shards. 
}
    \label{fig:data_sharding1}
\end{figure}

\textit{Data partition into small portions}. 
As shown in Fig.\ref{fig:data_sharding1},
each local dataset is partitioned into data shards. 
Each shard has a model and the final output is the aggregation of models from these shards. 
During the local training process, each data shard is independently trained and retains the model weight $\omega_{c,i}$ ($i \in [1, \tau]$). Ultimately, all models saved from the data shards are aggregated to construct the user's local model parameter $\omega_{c}$. 
The above description is formalized in Equation \ref{eq:locmodelparameter}, where $|D_i^c|$ represents the size of the $i$th shard and $|D^c|$ denotes the size of the user local data. 

\begin{equation}
\label{eq:locmodelparameter}
    \omega_c^t=\sum_{i=1}^{\tau}{\frac{\left|D_i^c\right|}{\left|D^c\right|}\omega_{c,i}^t}
\end{equation}
When a deletion request is made by the user,
it is only required to retrain the local model with the remaining dataset by removing the shard that contains the removed data. 
As a result, instead of re-initialization of user local model weights, it starts training from the checkpoint, which saves the time of retraining. The checkpoint is computed using the models from data shards that do not contain removed data as Equation \ref{equation:checkpoint}, where the $i$th shard is removed. 

\begin{equation}
    \omega_c^t=\sum_{j \neq i}^{\tau-1}{\frac{\left|D_j^c\right|}{\left|D^c\right|}\omega_{c,j}^t}
    \label{equation:checkpoint}
\end{equation}

After retraining, we delete the removed data in data shard $D_i^c$ and obtain the weights of the shard $D_i^c$ by subtracting the weights of other shards. 
As shown in Equation \ref{eq:lmw}, 
the new model weight of shard $D_i^c$ is computed by subtracting the model weights of other shards. 

\begin{equation}
\label{eq:lmw}
   \omega_{c,i}^t=\frac{|D^c|}{|D_i^c|}(\omega_c^{t+1}-\sum_{j \neq i}^{\tau-1}\frac{\left|D_j^c\right|}{\left|D^c\right|}\omega_{c,j}^t)
\end{equation}

\begin{figure}
    \centering
    \includegraphics[width=0.4\textwidth]{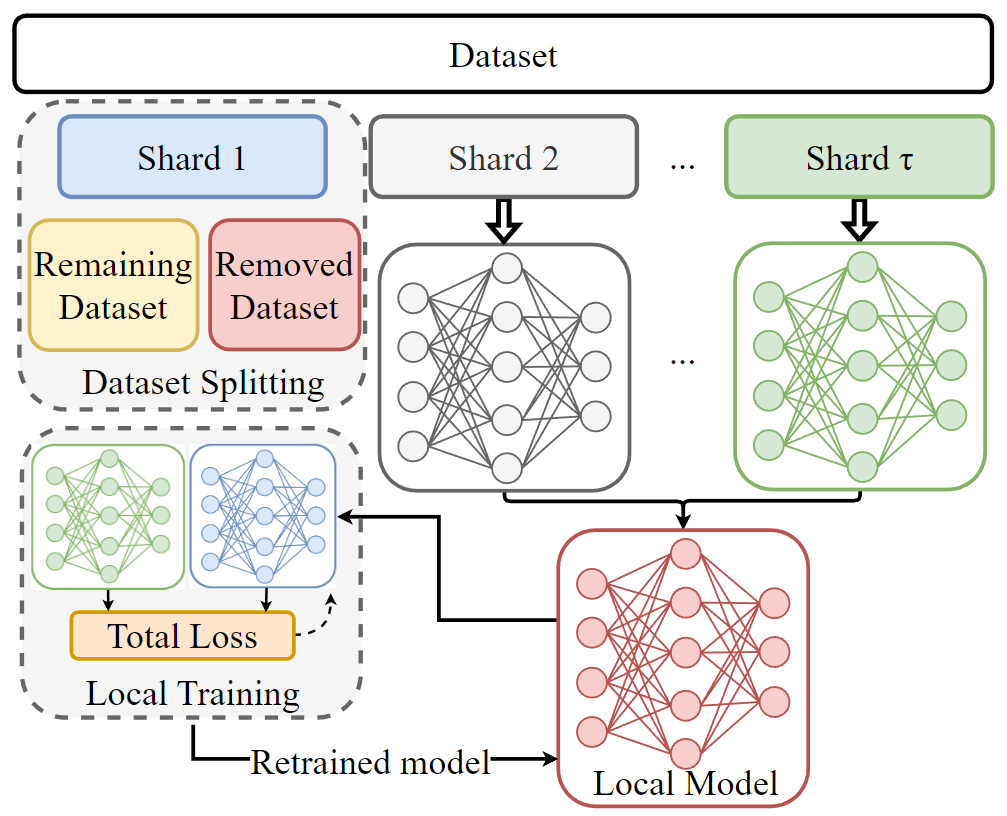}
    \caption{Retraining under data sharding.  In Shard 1, only partial data of the shard is deleted, it is required to retrain the model of the shard before model aggregation. }
    \label{fig:data_sharding2}
\end{figure}

In case, as shown in Fig.\ref{fig:data_sharding2},  if only partial data of a shard is deleted, it is required to retrain the model of the shard. 
If more than one shard is involved, the training of the multiple shards can be {parallelized}. 


\noindent\textbf{Extension}. 
We propose two mechanisms within the extension module. 
The first mechanism is applied to address the heterogeneity of user local data through adaptive distillation temperature. 
The second mechanism is employed to tackle the different quality levels of uploaded models via adaptive weights.

\textit{Adaptive distillation temperature proposed for addressing the heterogeneity of user local data}.
During the phase of model training, the models trained on clients with larger and more evenly distributed datasets behave better.
However, in real-life scenarios, 
the local datasets vary from {user to user}.  
For example,
{the sizes of local datasets may vary significantly among different users.}
It is natural to put more emphasis on models that exhibit better performance during model aggregation.



We formulate the above concept by using Equation \ref{eq:huld}, 
where $T_0$ is the initial distillation temperature, and $\alpha$ is the adjustment factor.
The principle behind the equation is that the amount of information decoupled by the student model from the outputs of the teacher model is directly proportional to the distillation temperature.
A higher distillation temperature can smooth the distribution of the teacher model's predicted results. 
When $T \le 1$, soft labels degrade into hard labels, and the distillation loss becomes equivalent to the traditional cross-entropy loss. We adjust the value of $T$ based on the sizes of the removed dataset and the remaining dataset.

\begin{equation}
\label{eq:huld}
    T=\alpha T_0 exp{\left(\frac{\left|D_r^c\right|}{\left|D_r^c\right|+\left|D_f^c\right|}\right)}
\end{equation}

\textit{Adaptive weight for addressing variety in the quality of uploaded models}. 
We use the Mean Squared Error (MSE) to measure the performance of different user local models and assign higher weights to models with higher prediction accuracy based on MSE. 
This approach ensures that the global model retains higher testing performance during model aggregation.
\begin{equation}
    \label{eq:mseweight}
    \mathcal{W}_{c} = \exp{\left(-\frac{{me}_{c}^t-\frac{1}{|C|} \sum_{i=1}^C \left({me_{ i}}^{t}\right)}{ \frac{1}{|C|} \sum_{i=1}^C \left({me_{i}}^{t}\right)}\right)}
\end{equation}
The detail is shown in Equation \ref{eq:mseweight}, where 
$\mathcal{W}_{c}$ denotes the weight allocated to the local model of the client $c$, 
${me}_{c}^t$ represents the MSE obtained by the client $c$ when testing its local model on the test set at the central server, $\frac{1}{|C|} \sum_{i=1}^C \left({me_{i}}^{t}\right)$ denotes the average of all clients' MSEs in the $t$ round of FL. Based on the definition of the client's weight, in the $t+1$th epoch,  the finally aggregated model parameter is defined in Equation \ref{eq:fullweight}, where $\theta$ is the normalization factor, i.e., $\theta = \sum_{c=1}^{C} \mathcal{W}_{c}$.

\begin{equation}
 \label{eq:fullweight}
    \omega^{t+1}=\frac{1}{\theta}\sum_{c=1}^C\mathcal{W}_{c}\omega_c^{t+1}
\end{equation}


\section{Performance of Goldfish}

In this section, 
we utilize different datasets and model architectures to evaluate the performance of the proposed forgetting approach. 
Before delving into the details of the experiment, we provide an overview of the setting below.

\subsection{Experimental Setup}

\textbf{Dataset Description}.
In the experiments, we utilized {four} public known ML datasets: the MNIST \cite{lecun1998gradient}, Fashion-MNIST(FMNIST)\cite{xiao2017fashion}, CIFAR-10 \cite{krizhevsky2009learning}, {and CIFAR-100 \cite{krizhevsky2009learning}}. 
As illustrated in Table \ref{tab:dataset}, these datasets encompass diverse attributes, dimensions, and numbers of classes.  
In the setting of the FL environment, we uniformly assigned the data from the {four} training datasets to all clients.  

\begin{table}[]
\centering
\caption{Dataset Description}
\label{tab:dataset}
\begin{tabular}{ccccc}
\hline
Dataset       & Dimensions & Classes & Training & Test  \\ \hline
MNIST         & 784        & 10      & 60000    & 10000 \\
Fashion-MNIST & 784        & 10      & 60000    & 10000 \\
CIFAR-10      & 3072       & 10      & 50000    & 10000 \\ 
{CIFAR-100}     & {3072}       & {100}     & {50000}    & {10000} \\ \hline
\end{tabular}
\end{table}

\textbf{Models}.
To enable the variety of models, we adopted {four} different models. 
In particular, the model for MNIST and FMNIST is a traditional LeNet-5 model \cite{liu2022right} \cite{lecun1998gradient} 
 consists of 2 convolutional layers, 2 max pool layers, and 2 fully connected layers in the end for prediction output. The {models} for CIFAR-10 are a  modified LeNet-5 consisting of 2 convolutional layers, 2 max pool layers, and 3 fully connected layers in the end for prediction output and {ResNet32, which is a variant of the Residual Network (ResNet) architecture \cite{he2016deep}. This model consists of 32 layers, including multiple residual blocks, each designed to learn residual functions with reference to the layer inputs.}
 {The model for CIFAR-100 is ResNet56, which is another variant of the Residual Network (ResNet) architecture \cite{he2016deep}. This model is designed to handle the increased complexity of the CIFAR-100 dataset. ResNet56 consists of 56 layers, including multiple residual blocks that are instrumental in enabling the network to learn deeper representations without the issues of vanishing gradients that can plague very deep networks.}
 All models are implemented using Pytorch, and experiments are done on a machine with one NVIDIA 1660ti GPU and a machine with one NVIDIA 3060 GPU.


\textbf{Hyperparameters}.
In the experiment,  the total number $C$ of clients is set to  $5$, $15$, and $25$. 
In addition, the batch size $B$ is set to $100$, and the learning rate $\eta$ is set to $0.001$. 
For the deletion rate, following \cite{wang2023machine},  we additionally include the values of $2$\%, $4$\%, $6$\%, $8$\%, $10$\%, and $12$\%. 
The momentum parameter $\beta$ is set to $0.9$ and the total number $N$ of shards, 
  following  \cite{bourtoule2021machine},   is set to  $1$, $3$, $6$, $9$, $12$, $15$, and $18$.

\textbf{Evaluation Metrics}. 
In the experiments, we evaluate the proposed approach in training speed, 
accuracy of prediction, unlearning ability, and robustness. 
In particular, following the validation approach in \cite{wu2022federated},  
we use backdoor attack to evaluate the unlearning ability. 
Additionally, we employ L2 distance, Jensen-Shannon Divergence (JSD), and T-test to evaluate the model running result.

\textbf{Baselines}.
To validate the effectiveness of the proposed approach, three baselines are set. 
The first baseline is an approach to retrain the model from the scratch \cite{zhang2023fedrecovery}, denoted as $B1$.
The second baseline is the rapid retraining method \cite{liu2022right}, denoted as $B2$. 
The third baseline is the federated unlearning algorithm \cite{chundawat2023can}, denoted as $B3$.

\subsection{Experiment}
The experiments are crafted to assess the effectiveness and forgetfulness of the proposed approaches. 



    

\begin{figure*}[h]
\label{fig:acc}
\centering
    \subfloat[]{\label{fig:mnist1}\includegraphics[width=0.32\textwidth]{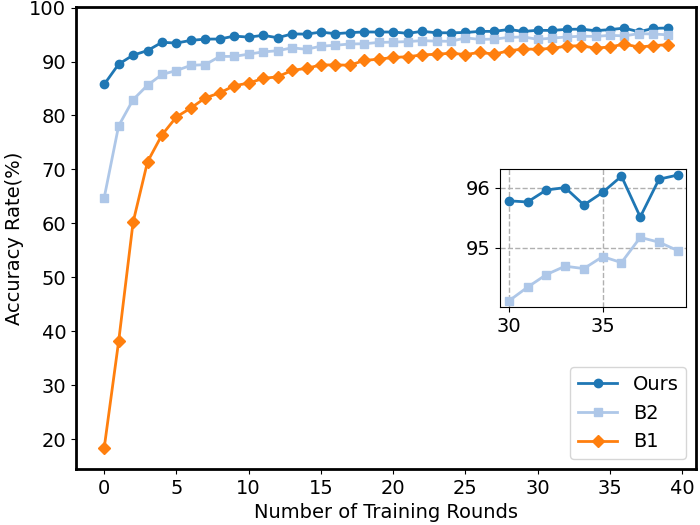}}
    \subfloat[]{\label{fig:fmnist1}\includegraphics[width=0.32\textwidth]{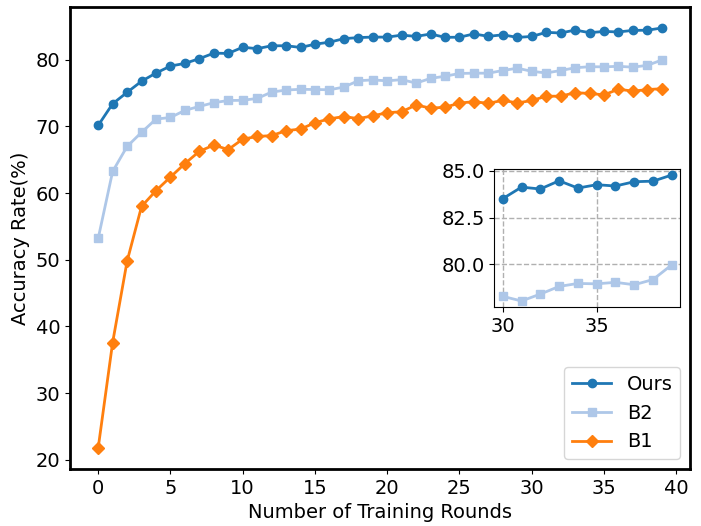}}
    \subfloat[]{\label{fig:cifar1}\includegraphics[width=0.32\textwidth]{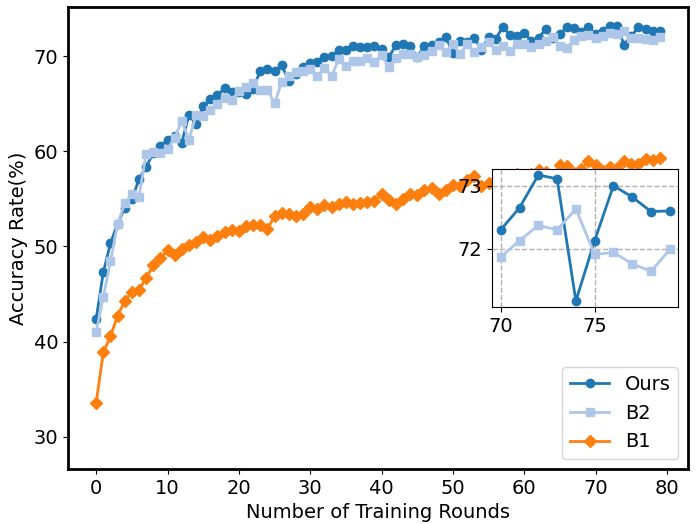}}
    \vspace{0.2cm} 
    \subfloat[]{\label{fig:cifar10-resnet32}\includegraphics[width=0.32\textwidth]{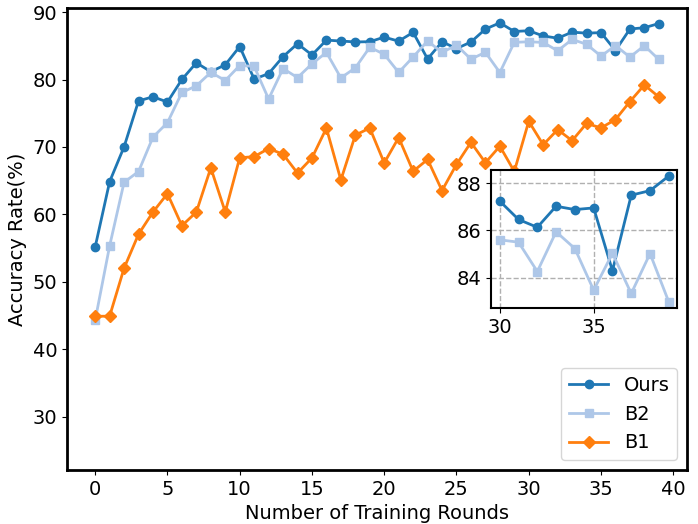}}
    \subfloat[]{\label{fig:cifar100-resnet56}\includegraphics[width=0.32\textwidth]{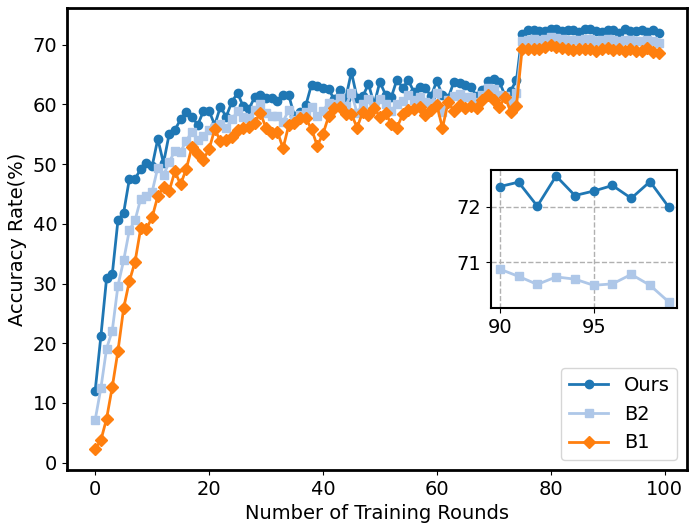}}
 \caption{Accuracy rate of (a) LeNet-5 model trained on the MNIST dataset, (b) LeNet-5 model trained on FMNIST dataset,  {(c) Modified LeNet-5 model trained on CIFAR-10 dataset, (d) ResNet32 model trained on CIFAR-10 dataset, and (e) ResNet56 model trained on CIFAR-100 dataset.} }
\end{figure*}

We first execute the retraining algorithms across four distinct datasets. 
Specifically, we set the batch size $B$ to $100$, and learning rate $\eta$ to $0.001$. 
In addition, {we employed models with accuracies of 94.5\%, 81.3\%, and 72.6\% as teacher models for the datasets MNIST, FMNIST, and CIFAR-100, respectively. Additionally, a modified LeNet-5 model with an accuracy of 72.6\% and a ResNet32 model with an accuracy of 92.6\% were utilized as teacher models for the CIFAR-10 dataset.}. 
The retraining results for {MNIST, FMNIST, CIFAR-10 and CIFAR-100} are illustrated in {Fig.\ref{fig:mnist1}, Fig.\ref{fig:fmnist1}, Fig.\ref{fig:cifar1}, Fig.\ref{fig:cifar10-resnet32}, and Fig.\ref{fig:cifar100-resnet56}} respectively.
Upon comparison, it becomes evident that our approach attains the highest accuracy, followed by $B_2$ in second place, while $B_1$ exhibits the lowest accuracy.



We examine the success rate of backdoor attacks on models across various deletion rate settings. The objective is to assess whether our proposed approach effectively prevents the retention of pertinent information about backdoor samples in the model after retraining.
In our experiments, we compare our proposed method with the original model, $B1$ and $B3$.
Following the configuration of  \cite{zhu2023heterogeneous}, we set the distillation temperature $T$ to $3$, the weight factor $\mu_d$ to $1.0$, and $\mu_c$ to $0.25$.
{ B2 speeds up the retraining process by preserving historical gradients and other information, which is the same as B1. Both retrain from scratch. Therefore, it is not included here. }

\begin{figure*}[h]
\label{fig:backdoor}
\centering
    \subfloat[]{\label{fig:mnist2}\includegraphics[width=0.32\textwidth]{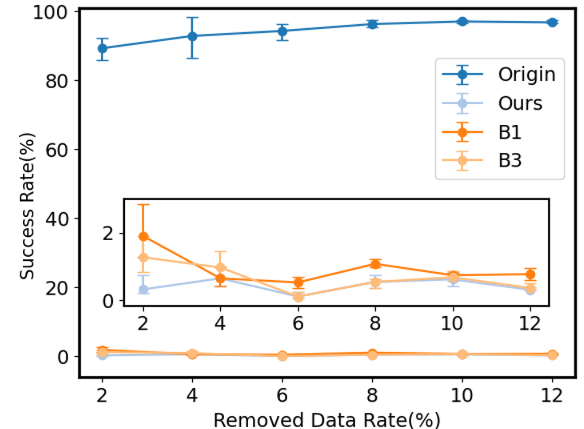}}
    \subfloat[]{\label{fig:fmnist2}\includegraphics[width=0.32\textwidth]{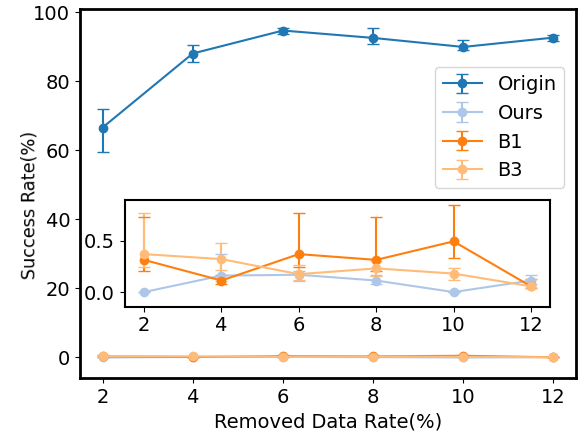}}
    \subfloat[]{\label{fig:cifar2}\includegraphics[width=0.32\textwidth]{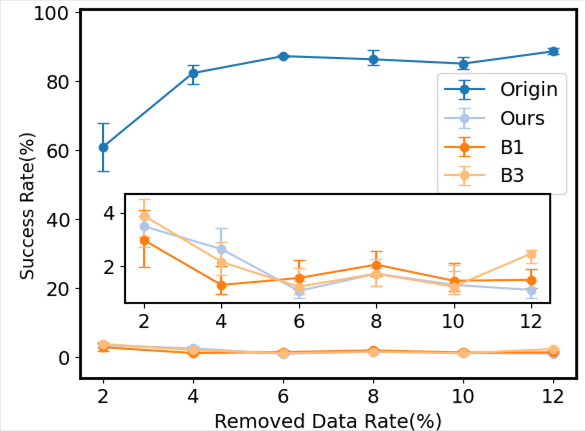}}
    \vspace{0.2cm} 
    \subfloat[]{\label{fig:cifar10-resnet32-backdoor}\includegraphics[width=0.32\textwidth]{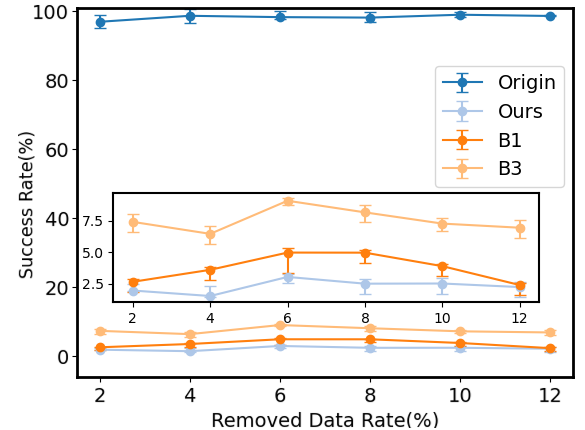}}
    \subfloat[]{\label{fig:cifar100-resnet56-backdoor}\includegraphics[width=0.32\textwidth]{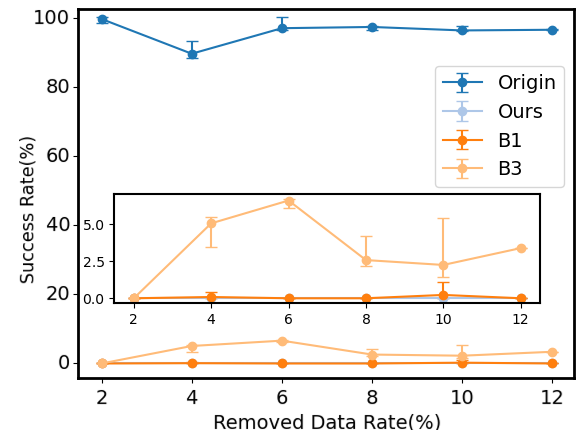}}
 \caption{Success Rate of backdoor attack of models under different removed data rates on the (a) MNIST dataset, (b) FMNIST dataset,  {(c) CIFAR-10 dataset, (d) CIFAR-10 dataset, and (e) CIFAR-100 dataset.}}
\end{figure*} 

The experimental result is shown in 
{Fig.\ref{fig:mnist2}, Fig.\ref{fig:fmnist2}, Fig.\ref{fig:cifar2}, Fig.\ref{fig:cifar10-resnet32-backdoor}, and Fig.\ref{fig:cifar100-resnet56-backdoor}.}
From the figures, 
we can learn that the original model consistently maintains a high backdoor attack success rate, indicating contamination of model parameters and the emergence of specific backdoor prediction patterns.
In contrast, our method, as well as $B1$ and $B3$, consistently shows low backdoor attack success rates, particularly on MNIST and FMNIST datasets. 
Our method consistently maintains the lowest backdoor attack success rate across different deletion rates with minimal accuracy error. 
This underscores the effective prevention of information retention from the removed set during retraining.
\begin{table*}[]
\caption{Accuracy rate and success rate of backdoor attack of models on the MNIST dataset.}
\label{tabel:MNIST acc and backdoor}
\centering
\begin{tabular}{ccccccccc}
\hline
\multirow{2}{*}{Deleted data Rate} & \multicolumn{2}{c}{origin} & \multicolumn{2}{c}{Ours} & \multicolumn{2}{c}{$B1$} & \multicolumn{2}{c}{$B3$} \\ \cline{2-9} 
                                   & acc        & backdoor      & acc       & backdoor     & acc          & backdoor       & acc          & backdoor       \\ \hline
2                                  & 85.67      & 89.20          & 92.67     & 0.31         & 95.00           & 1.90            & 91.33        & 1.27           \\
4                                  & 86.58      & 92.79         & 92.58     & 0.32         & 95.50         & 0.65           & 90.67        & 1.29           \\
6                                  & 85.94      & 94.24         & 95.06     & 0.10          & 95.94        & 0.52           & 93.61        & 0.21           \\
8                                  & 86.08      & 96.23         & 94.25     & 0.54         & 95.38        & 1.08           & 91.18        & 0.77           \\
10                                 & 85.63      & 96.98         & 95.10      & 0.68         & 95.63        & 0.74           & 95.33        & 0.74           \\
12                                 & 85.33      & 96.71         & 94.75     & 0.41         & 95.36        & 0.77           & 95.56        & 0.41           \\ \hline
\end{tabular}
\end{table*}

\begin{table*}[]
\caption{Accuracy rate and success rate of backdoor attack of models on the FMNIST dataset.}
\label{tabel:FMNIST acc and backdoor}
\centering 
\begin{tabular}{ccccccccc}
\hline
\multirow{2}{*}{Deleted data Rate} & \multicolumn{2}{c}{origin} & \multicolumn{2}{c}{Ours} & \multicolumn{2}{c}{$B1$} & \multicolumn{2}{c}{$B3$} \\ \cline{2-9} 
                                   & acc        & backdoor      & acc       & backdoor     & acc          & backdoor       & acc          & backdoor       \\ \hline
2                                  & 75.33      & 66.56         & 78.67     & 0.00         & 79.50        & 0.47           & 73.50        & 0.31           \\
4                                  & 72.50      & 88.00         & 76.83     & 0.16         & 80.33        & 0.31           & 75.17        & 0.16           \\
6                                  & 72.89      & 94.65         & 77.17     & 0.17         & 80.50        & 0.00           & 75.61        & 0.00           \\
8                                  & 71.50      & 92.52         & 77.33     & 0.11         & 81.33        & 0.34           & 74.62        & 0.17           \\
10                                 & 71.90      & 89.89         & 78.57     & 0.00         & 81.33        & 0.06           & 74.13        & 0.07           \\
12                                 & 71.17      & 92.60         & 77.58     & 0.11         & 78.17        & 0.06           & 75.08        & 0.05           \\ \hline
\end{tabular}
\end{table*}

\begin{table*}[]
\centering 
\caption{ Accuracy rate and success rate of backdoor attack of models on the CIFAR-10 dataset.}
\label{tabel:CIFAR10 acc and backdoor}
\begin{tabular}{ccccccccc}
\hline
\multirow{2}{*}{Deleted data Rate} & \multicolumn{2}{c}{origin} & \multicolumn{2}{c}{Ours} & \multicolumn{2}{c}{$B1$} & \multicolumn{2}{c}{$B3$} \\ \cline{2-9} 
                       & acc          & backdoor        & acc         & backdoor       & acc        & backdoor      & acc        & backdoor      \\ \hline
2                      & 78.54        & 97.14       & 87.11       & 1.90       & 84.61      & 2.86      & 87.53      & 7.62      \\
4                      & 77.57        & 99.05       & 86.31       & 1.43       & 85.13      & 3.81      & 85.64      & 6.67      \\
6                      & 75.47        & 98.10       & 86.28       & 3.16       & 84.69      & 5.38      & 87.69      & 9.18      \\
8                      & 78.65        & 98.28       & 86.40       & 2.71       & 84.97      & 5.17      & 88.72      & 8.37      \\
10                     & 78.52        & 99.02       & 86.21       & 2.73       & 84.96      & 4.10      & 87.71      & 7.42      \\
12                     & 78.52        & 98.54       & 86.31       & 2.44       & 84.97      & 2.60      & 86.64      & 7.15      \\ \hline
\end{tabular}
\end{table*}

\begin{table*}[]
\centering 
\caption{ Accuracy rate and success rate of backdoor attack of models on the CIFAR-100 dataset.}
\label{tabel:CIFAR100 acc and backdoor}
\begin{tabular}{ccccccccc}
\hline
\multirow{2}{*}{Deleted data Rate} & \multicolumn{2}{c}{origin} & \multicolumn{2}{c}{Ours} & \multicolumn{2}{c}{$B1$} & \multicolumn{2}{c}{$B3$} \\ \cline{2-9} 
                        & acc         & backdoor         & acc         & backdoor       & acc        & backdoor      & acc        & backdoor      \\ \hline
2                       & 57.83       & 100.00       & 64.29       & 0          & 59.22      & 0         & 65.22      & 0.00      \\
4                       & 55.14       & 89.47        & 65.22       & 0          & 55.67      & 0         & 62.24      & 5.26      \\
6                       & 51.04       & 96.67        & 65.31       & 0          & 57.84      & 0         & 65.24      & 6.67      \\
8                       & 53.96       & 97.62        & 64.39       & 0          & 56.91      & 0         & 63.08      & 2.38      \\
10                      & 50.96       & 96.30        & 65.08       & 0          & 55.67      & 0         & 65.00      & 1.85      \\
12                      & 51.26       & 96.61        & 66.44       & 0          & 55.71      & 0         & 67.18      & 3.39      \\ \hline
\end{tabular}
\end{table*}

\begin{table}[]
\caption{Evaluation based on JSD, L2, and T-test on the MNIST dataset.}
\label{tabel:MNIST metrics}
\centering
\begin{tabular}{ccccccc}
\hline
\multirow{2}{*}{Deleted data Rate} & \multicolumn{3}{c}{$B3$} & \multicolumn{3}{c}{Ours} \\ \cline{2-7} 
                                   & JSD     & L2      & T-test    & JSD    & L2    & T-test  \\ \hline
2                                  & 0.69    & 0.49    & 0.95      & 0.56   & 0.14  & 0.29    \\
4                                  & 0.69    & 0.27    & 0.67      & 0.69   & 0.15  & 0.45    \\
6                                  & 0.69    & 0.25    & 0.74      & 0.63   & 0.06  & 0.33    \\
8                                  & 0.63    & 0.27    & 0.85      & 0.63   & 0.07  & 0.43    \\
10                                 & 0.69    & 0.06    & 0.31      & 0.63   & 0.07  & 0.35    \\
12                                 & 0.69    & 0.06    & 0.28      & 0.54   & 0.07  & 0.35    \\ \hline
\end{tabular}
\end{table}

\begin{table}[]
\caption{Evaluation based on JSD, L2, and T-test on the FMNIST dataset.}
\label{tabel:FMNIST metrics}
\centering
\begin{tabular}{ccccccc}
\hline
\multirow{2}{*}{Deleted data Rate} & \multicolumn{3}{c}{$B3$} & \multicolumn{3}{c}{Ours} \\ \cline{2-7} 
                                   & JSD     & L2      & T-test    & JSD    & L2    & T-test  \\ \hline
2                                  & 0.63    & 0.25    & 0.92      & 0.63   & 0.33  & 0.82    \\
4                                  & 0.69    & 0.27    & 0.81      & 0.69   & 0.32  & 0.59    \\
6                                  & 0.69    & 0.3     & 0.71      & 0.69   & 0.31  & 0.67    \\
8                                  & 0.69    & 0.18    & 0.82      & 0.69   & 0.21  & 0.64    \\
10                                 & 0.69    & 0.29    & 0.69      & 0.69   & 0.22  & 0.53    \\
12                                 & 0.69    & 0.22    & 0.69      & 0.69   & 0.16  & 0.58    \\ \hline
\end{tabular}
\end{table}

\begin{table}[]
\centering
\caption{Evaluation based on JSD, L2, and T-test on the CIFAR-10 dataset.}
\label{tabel:CIFAR-10 metrics}
\begin{tabular}{ccccccc}
\hline
\multirow{2}{*}{Deleted data Rate} & \multicolumn{3}{c}{$B3$} & \multicolumn{3}{c}{Ours} \\ \cline{2-7} 
                                   & JSD     & L2      & T-test    & JSD    & L2    & T-test  \\ \hline
2                                  & 0.70    & 0.35    & 0.82      & 0.70   & 0.34  & 0.71    \\
4                                  & 0.63    & 0.36    & 0.85      & 0.63   & 0.37  & 0.73    \\
6                                  & 0.63    & 0.36    & 0.75      & 0.63   & 0.38  & 0.67    \\
8                                  & 0.59    & 0.23    & 0.81      & 0.58   & 0.26  & 0.74    \\
10                                 & 0.67    & 0.37    & 0.79      & 0.66   & 0.33  & 0.63    \\
12                                 & 0.63    & 0.22    & 0.73      & 0.63   & 0.17  & 0.68    \\ \hline
\end{tabular}
\end{table}


In  
{Table \ref{tabel:MNIST acc and backdoor}, Table \ref{tabel:FMNIST acc and backdoor}, Table \ref{tabel:CIFAR10 acc and backdoor}, and Table \ref{tabel:CIFAR100 acc and backdoor}},
we present 
the result of running different models on the test set with various deletion rates. 
From the tables, it is evident that our proposed method consistently maintains high accuracy across different deletion rates. This indicates the effectiveness of our approach in ensuring forgetting while preserving predictive accuracy on the test set. 

As demonstrated in Table \ref{tabel:MNIST metrics}, Table \ref{tabel:FMNIST metrics}, and Table \ref{tabel:CIFAR-10 metrics}, we compare $B3$ and our approach with $B1$ in terms of JSD and L2 distance
{on MNIST, FMNIST and CIFAR-10 datesets}. Additionally, we conduct a T-test to compare $B3$ and our approach with the original model.
{Jensen-Shannon Divergence (JSD) and L2 Distance are statistical metrics used to quantify the dissimilarity between two probability distributions. JSD measures the average Kullback-Leibler Divergence between the two distributions, whereas L2 Distance measures the mean squared error between them.}
A small JSD and L2 distance implies a high similarity between two distributions. 
{
A T-test is a statistical method used to determine if there is a significant difference between the means of two groups.
In a T-test, the p-value (probability value) indicates the probability of obtaining the observed results (or more extreme) if the null hypothesis (\textit{i.e.}, the means of two groups are equal) is true. 
 The smaller the p-value, the stronger the evidence against the null hypothesis, indicating a greater likelihood of a significant difference between the means of the two samples.
}
From the tables, we can learn that both our approach and $B3$ have small L2 values. Moreover, our method has a smaller JSD value compared with $B3$, indicating that the predictive results obtained through our method are closer to those obtained through $B1$.
In the context of the T-test, our algorithm consistently yields smaller $p$-values in most cases. This suggests significant differences between the predictive patterns obtained through our algorithm and those generated by backdoor attacks.

{In order to investigate the importance of different components of the loss function, following \cite{zhu2023heterogeneous}, we conducted an ablation experiment on the loss function. We trained the ResNet32 model on the CIFAR-10 dataset, considering four combinations of loss functions: hard loss only, without distillation loss (\textit{i.e.}, hard loss and confusion loss),  without confusion loss (\textit{i.e.}, hard loss and distillation loss), and total loss (\textit{i.e.}, hard loss, distillation loss, and confusion loss) to study the roles of different parts of the loss function. The experimental result is shown in  Table \ref{tabel:ablation study of loss}.}

{
After analyzing the experimental data, we observe that the model trained without distillation loss exhibits a low success rate in backdoor attacks while compromising overall accuracy. However, the model trained without confusion loss maintains a high accuracy and success rate in backdoor attacks.
Notably, the incorporation of total loss ensures both elevated model accuracy and a reduced success rate in backdoor attacks. This underscores the advantageous role of distillation loss in expediting the training process and highlights the significance of confusion loss in facilitating more effective forgetting of backdoor data during training. Additionally, we note a marginal decrease in accuracy during the initial training stages for the model lacking confusion loss compared to other models. This decline is attributed to the absence of confusion loss, leading the model to adopt incorrect predictive patterns.}

{Furthermore, we have introduced additional loss functions, namely Focal loss \cite{Lin_2017_ICCV} and Negative Log Likelihood (NLL) loss \cite{lecun1998gradient}, in addition to the traditional cross-entropy loss employed as the hard loss.
This sequential inclusion aims to systematically assess the 
the compatibility of our framework with various types of loss functions, as illustrated in the experimental results presented in Table \ref{tabel:ablation study of different loss functions}. We denote the scenario where the cross-entropy loss function is utilized as the hard loss as  `Total loss $\alpha$', while the scenarios employing Focal loss and NLL loss are denoted as  `Total loss $\beta$' and  `Total loss $\gamma$', respectively. The table demonstrates consistently high accuracy in testing across different loss functions, coupled with a consistently low success rate in backdoor attacks. This underscores the robust compatibility of the proposed framework with diverse loss functions.}


\begin{table*}[]
\centering 
\caption{ Ablation Study of the Importance of Different Components of the Loss Function}
\label{tabel:ablation study of loss}
\begin{tabular}{cccccc}
\hline
Epoch               & Metrics & Hard loss only & w/o Distillation loss & w/o Confussion loss & Total loss \\ \hline
\multirow{2}{*}{10} & acc     & 80.73          & 80.57                 & 78.91               & 83.75      \\
                    & backdoor    & 5.08           & 2.93                  & 6.45                & 3.71       \\
\multirow{2}{*}{20} & acc     & 84.63          & 84.19                 & 84.25               & 85.84      \\
                    & backdoor    & 5.08           & 4.88                  & 7.03                & 5.08       \\
\multirow{2}{*}{30} & acc     & 87.03          & 85.40                 & 86.80               & 88.22      \\
                    & backdoor    & 1.95           & 2.60                  & 3.13                & 1.10       \\
\multirow{2}{*}{40} & acc     & 86.61          & 86.63                 & 87.71               & 88.56      \\
                    & backdoor    & 5.08           & 2.93                  & 2.73                & 2.44       \\ \hline
\end{tabular}
\end{table*}

\begin{table}[]
\centering 
\caption{ Compatibility Study of Different Loss Functions.}
\label{tabel:ablation study of different loss functions}
\begin{tabular}{ccccc}
\hline
Epoch               & Metrics  & Total loss $\alpha$ & Total loss $\beta$ & Total loss $\gamma$ \\ \hline
\multirow{2}{*}{10} & acc      & 83.75               & 83.37              & 83.71               \\
                    & backdoor & 3.71                & 3.81               & 2.86                \\
\multirow{2}{*}{20} & acc      & 85.84               & 88.03              & 84.63               \\
                    & backdoor & 5.08                & 1.9                & 1.95                \\
\multirow{2}{*}{30} & acc      & 88.22               & 87.54              & 87.71               \\
                    & backdoor & 1.10                & 1.9                & 2.86                \\
\multirow{2}{*}{40} & acc      & 88.56               & 88.49              & 87.92               \\
                    & backdoor & 2.44                & 0.95               & 2.71                \\ \hline
\end{tabular}
\end{table}

To measure the efficacy of data sharding, we seek answers to two pivotal questions:
\begin{itemize}
    \item Does the model exhibit effective convergence with the application of data sharding?
    \item Does the utilization of data sharding contribute to an expedited retraining process?
\end{itemize}

\begin{figure}
    \centering
    \includegraphics[width=0.45\textwidth]{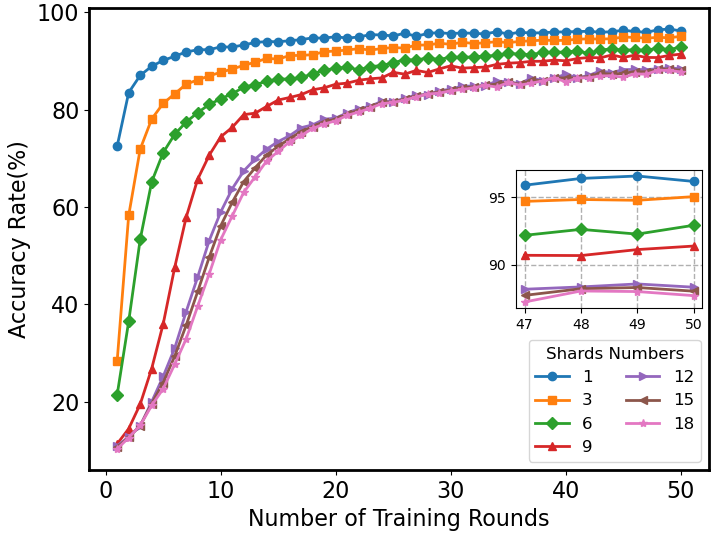}
    \caption{
    Accuracy rate of models trained on the MNIST dataset with various numbers of shards.}
    \label{fig:p1}
\end{figure}

Following  the experimental configuration \cite{bourtoule2021machine}, 
the number of shards is set to $1$, $3$, $6$, $9$, $12$, $15$, and $18$. 
In addition, the MNIST dataset is employed.  
Initially, each data shard is assigned with a model and each of them is trained over the shard. 
After that, those models are locally aggregated and the aggregated model becomes the client's local model. 
The training result is shown in Fig.\ref{fig:p1}. 
From the figure, it is evident that the rate of accuracy improvement decelerates with an increasing number of data shards. This phenomenon is attributed to the partiality of data within each shard, resulting in the model being biased towards local data. However, irrespective of the number of shards, the convergence trend of model accuracy remains consistent.

{
Moreover, for the optimal selection of the number of data shards, it depends on the acceptable accuracy loss and benefits from a reduced number of training rounds in data deletion. In formal, we could write the above problem as computing the maximum value of $\left(rr\cdot c_1-al\cdot c_2\right)$, where $rr$ is the reduced number of training rounds, $al$ is the accuracy loss, $c_1$ is the benefit of one reduced training round, and $c_2$ is the cost of one percent of accuracy loss. $rr$ and $al$ are corresponding to the chosen of the number of data shards while $c_1$ and $c_2$ are set based on the user preference. }

\begin{figure*}[h]
\label{fig:acshard}
\centering
    \subfloat[]{\label{fig:p2}\includegraphics[width=0.33\textwidth]{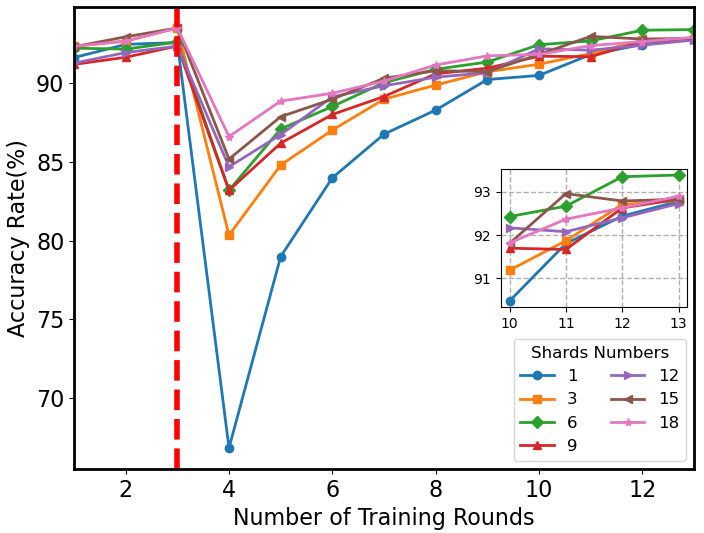}}
    \subfloat[]{\label{fig:p3}\includegraphics[width=0.33\textwidth]{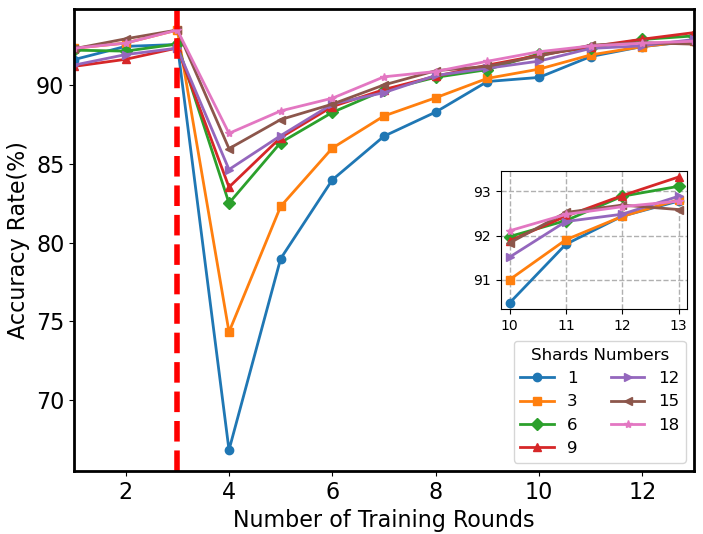}}
    \subfloat[]{\label{fig:p4}\includegraphics[width=0.33\textwidth]{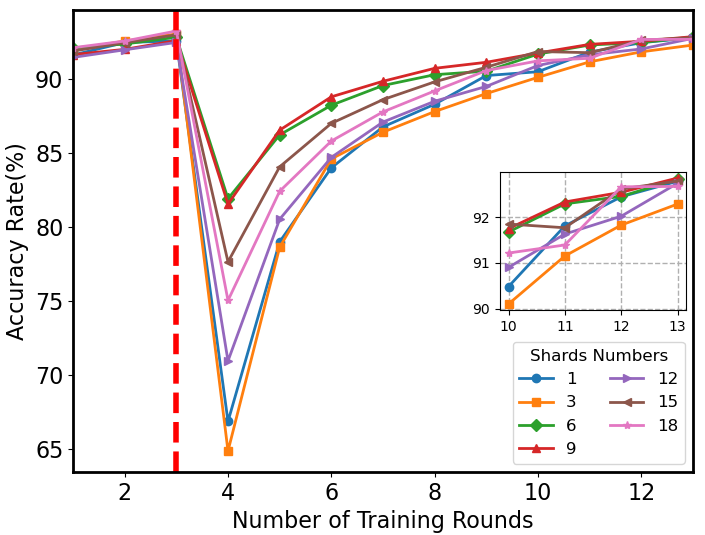}}
 \caption{
 Model Accuracy on MNIST Dataset with Different Data Shards Numbers at (a) 2\%, (b) 6\%, and (c) 10\% Deletion Rates.
 }
\end{figure*} 

\begin{figure*}[h]
\label{fig:acuneven}
\centering
    \subfloat[]{\label{fig:p6}\includegraphics[width=0.34\textwidth]{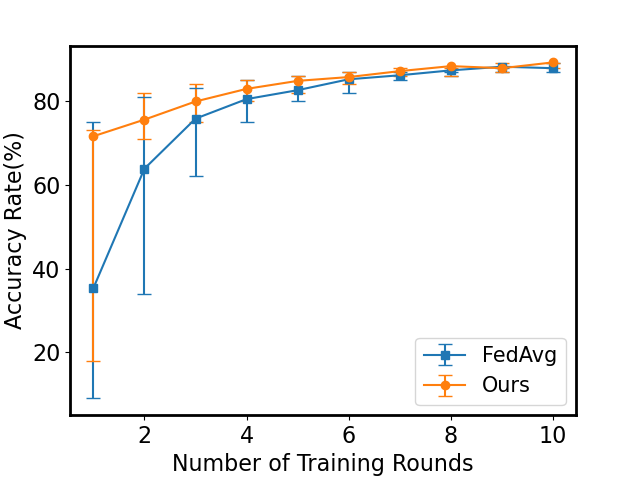}}
    \subfloat[]{\label{fig:p7}\includegraphics[width=0.34\textwidth]{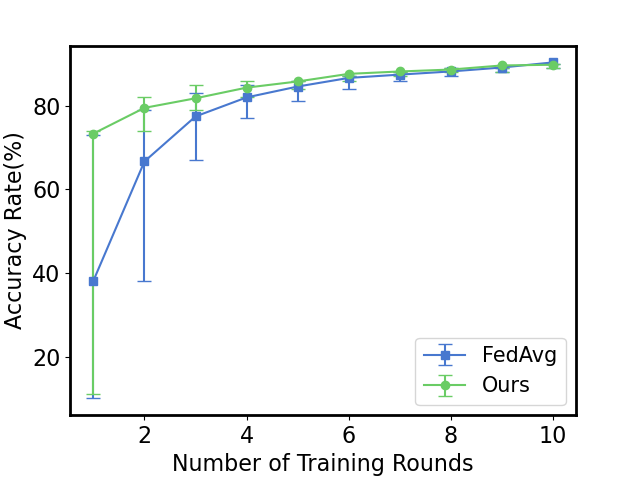}}
    \subfloat[]{\label{fig:p8}\includegraphics[width=0.34\textwidth]{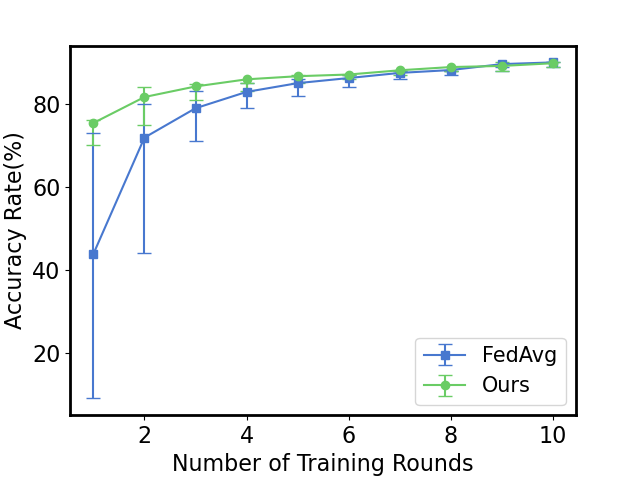}}
 \caption{
 Accuracy Comparison of FedAvg and Ours on MNIST with Uneven Local Data Distribution for (a) 5, (b) 15, and (c) 25 Clients.
 }
\end{figure*} 



To investigate the second question, we configured deletion rates at 2\%, 6\%, and 10\% and observed the variations in model accuracy before and after data deletion across different numbers of data shards. The experimental results are depicted in Fig.\ref{fig:p2}, Fig.\ref{fig:p3}, and Fig.\ref{fig:p4}, with the red dashed line at $3$ indicating the moment of the data deletion operation.
From Fig.\ref{fig:p2}, it is clear that when the deletion rate is 2\%, the model utilizing data sharding attains higher accuracy after the initial round of training following data deletion, compared to the model without data sharding (i.e., the number of shards is 1). 
Furthermore, it rapidly approaches the convergence accuracy value before data deletion. This occurs because, in scenarios with a low deletion rate, the deleted data only resides within one of the data shards, thus gaining an advantage over the model without data sharding at the beginning of training. 

At a 6\% deletion rate, illustrated in Fig.\ref{fig:p3}, the number of data shards containing deleted data increases. Consequently, models with fewer data shards experience a decrease in accuracy.
At a 10\% deletion rate, as depicted in Fig.\ref{fig:p4}, all models utilizing data sharding witness a decline in accuracy. The accuracy with three shards is similar to that with one shard because all shards necessitate retraining when data deletion occurs. However, when the number of shards is 6 and 9, the accuracy remains very high, indicating a trade-off between the number of shards and efficacy.
{Based on our empirical results, we recommend using a moderate number of shards that allows for efficient retraining without compromising the model's generalization ability. Specifically, we found that using 6 to 9 shards provided a good balance, as it maintained high accuracy while enabling the model to adapt quickly to data deletions.}

{Our findings indicate that although the initial advantage in convergence speed is not maintained throughout the entire training process, the data sharding approach shows its resilience against dataset change. Specifically, after experiencing the change in the dataset, the data sharding model can recover faster than a non-sharded model. 
}

We also evaluate the robustness of the proposed model aggregation approach. For benchmarking against the state-of-the-art, we compare our approach with FedAvg \cite{mcmahan2017communication}.
In this evaluation, we aim to answer the following two questions: 

\begin{enumerate}
    \item Does our proposed federated model aggregation approach exhibit comparable effectiveness to FedAvg when the client's local data is independently and identically distributed?
    \item Does our proposed federated model aggregation approach surpass FedAvg in preserving the performance capability of the model when the client's local data is highly heterogeneous?
\end{enumerate}

To explore the answers, 
we set the total number of clients to 5, 15, and 25, and randomly assign the data from MNIST dataset to all users equally. 

\begin{figure}
    \centering
    \includegraphics[width=0.45\textwidth]{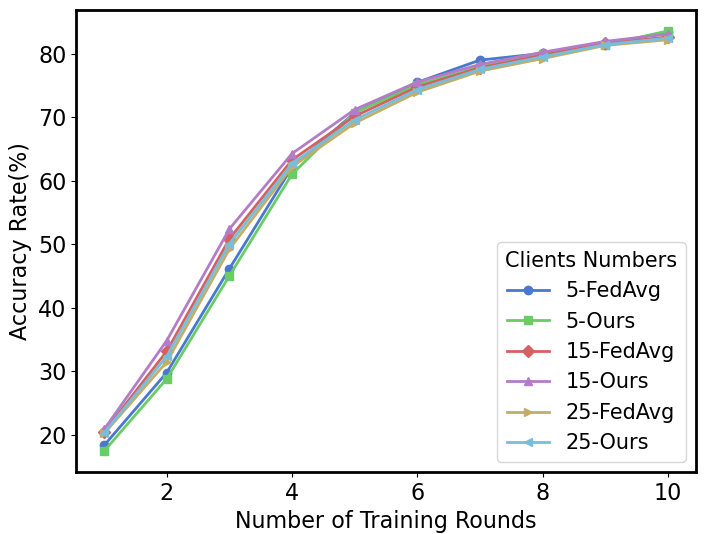}
    \caption{
    Accuracy Rate of FedAvg and Ours trained on the MNIST dataset with various numbers of clients. The local datasets for all users are independently and identically distributed.
    }
    \label{fig:p5}
\end{figure}

As shown in Fig.\ref{fig:p5}, the accuracy obtained by our aggregation method and FedAvg aggregation method exhibit virtually identical variations under the condition of uniform data distribution. This indicates comparable applicability for both methods when confronted with uniformly distributed data.

{
We define heterogeneity as the diversity in datasets across different users. 
To construct a heterogeneous dataset,  data is randomly assigned to each user. To quantitatively evaluate the level of heterogeneity, we calculate the variance of dataset sizes among users and analyze the range of maximum and minimum accuracy attained by independently training models on local datasets for all users. These metrics are presented in Table \ref{tabel:heterogeneity}.
}

\begin{table}[]
\caption{ Representation of Data Heterogeneity.}
\label{tabel:heterogeneity}
\centering
\begin{tabular}{cccc}
\hline
Clients Numbers & Variance        & Min acc & Max acc \\\hline
5               & 2.00$\times$$10^6$ & 9.3     & 75.6    \\
15              & 1.87$\times$$10^7$ & 10.1    & 73.3    \\
25              & 5.20$\times$$10^7$ & 9.8     & 73.4   \\ \hline
\end{tabular}
\end{table}

Before performing model aggregation, we assess the accuracy of local models on the test set and compute the accuracy of the global model. The resulting accuracy is plotted with error bars representing precision statistics for all users. This facilitates a comparison of the aggregation capabilities between FedAvg and our proposed method. Figures \ref{fig:p6}, \ref{fig:p7}, and \ref{fig:p8} show wide-ranging error bars of FedAvg at the initial stage due to the heterogeneity of local data among users. This heterogeneity leads to significant variations in the accuracy of local models during the initial training phase. 
However, our proposed federated model aggregation method effectively addresses the challenge of data heterogeneity in the initial training phase. We assign higher weights to local models demonstrating superior performance on the test set during the aggregation process. This strategic approach allows the global model to retain the parameter distribution of high-performing models, thereby enhancing overall performance and retraining efficiency.

{
\section{Discussion}

 In this section, we discuss the challenges and our future research directions of federated unlearning.  

The primary challenge lies in striking a balance between the efficiency and validity of federated unlearning. This process aims to expunge specific client data from the global model without compromising model performance. Successfully addressing this challenge demands innovative strategies to streamline the unlearning process while preserving user privacy and maintaining model accuracy.
Another challenge is to adapt to a diverse array of heterogeneous clients. In the dynamic landscape of federated unlearning, where clients may join or leave, and data distributions can vary, the federated unlearning scheme must exhibit both flexibility and resilience.

As part of our future research direction, we aim to enhance the accuracy rate of the current data sharding model. Additionally, we plan to explore strategies to effectively address the dynamism and heterogeneity inherent in client environments.


 
}

\section{Conclusion}
In this paper, we introduce {a} new paradigm of designing machine unlearning algorithms, named Goldfish.
Moreover, we instantiate each module of Goldfish to achieve better efficiency and validity by adopting knowledge distillation technique in basic model, introducing a novel loss function, proposing an optimization module, and illustrating an extension implementation. 
Furthermore, we conduct comprehensive experiments to validate the effectiveness of approaches.



\bibliographystyle{ieeetr}
\bibliography{reference}

\end{document}